\documentclass[10pt, a4paper]{article}

\usepackage{lrec-coling2024} 
\usepackage{tipa}
\usepackage{xspace}

\newcommand{\PRAUTOCAL}{\textsc{Prautocal}\xspace}
\newcommand{\ProjectCode}{\textbf{PID2021-126315OB-I00}\xspace}
\newcommand{\FundingAgency}{\textbf{MCIN / AEI / 10.13039/501100011033 / FEDER, EU}\xspace}

\title{A Comprehensive Rubric for Annotating Pathological Speech}

\name{Mario Corrales-Astorgano$^1$, David Escudero-Mancebo$^1$, Lourdes Aguilar$^2$, \\
{\bf \large Valle Flores-Lucas$^1$, Valentín Cardeñoso-Payo$^1$, Carlos Vivaracho-Pascual$^1$} \\ 
{\bf \large and César González-Ferreras$^1$}}

\address{Universidad de Valladolid$^1$, Universitat Autònoma de Barcelona$^2$ \\
mcorrales@infor.uva.es, descuder@infor.uva.es, Lourdes.Aguilar@uab.cat, \\ 
mariavalle.flores@uva.es, valentin.cardenoso@uva.es, cevp@infor.uva.es, 
cesargf@uva.es}

\abstract{Rubrics are a commonly used tool for labeling voice corpora in speech quality assessment, although their application in the context of pathological speech remains relatively limited. In this study, we introduce a comprehensive rubric based on various dimensions of speech quality, including phonetics, fluency, and prosody. The objective is to establish standardized criteria for identifying errors within the speech of individuals with Down syndrome, thereby enabling the development of automated assessment systems. To achieve this objective, we utilized the \PRAUTOCAL corpus. To assess the quality of annotations using our rubric, two experiments were conducted, focusing on phonetics and fluency. For phonetic evaluation, we employed the Goodness of Pronunciation (GoP) metric, utilizing automatic segmentation systems and correlating the results with evaluations conducted by a specialized speech therapist. While the obtained correlation values were not notably high, a positive trend was observed. In terms of fluency assessment, deep learning models like wav2vec were used to extract audio features, and we employed an SVM classifier trained on a corpus focused on identifying fluency issues to categorize \PRAUTOCAL corpus samples. The outcomes highlight the complexities of evaluating such phenomena, with variability depending on the specific type of disfluency detected.
\\ \newline 
 }

\begin{document}

\maketitleabstract

\section{Introduction}

The use of rubrics to assess speech quality is common in domains related
to education, where it is necessary to determine the level attained by a
particular speaker in acquiring specific competencies. In reading proficiency
or L2 pronunciation training, specific rubrics have been developed to
identify students' deficiencies in their educational process 
(see Section \ref{sec:state}).
The application of these rubrics is contributing to the development of automated systems that assess pronunciation quality, as demonstrated in studies like 
\citet{bailly2022automatic} and \citet{prabhu2022cnn}.

In the field of pathological speech, progress towards automatic systems for
identifying specific problems is not advancing as rapidly for various
reasons. One of them is that it is not enough to simply annotate audio with a
quality level that assesses the degree of nativeness or fluency of speech. In
the case of pathological speech, it is important to diagnose the specific
problem that the speaker has (from a long list of potential issues, such as
dysarthria, stuttering, cluttering, \ldots) or the symptom that
indicates a possible pathology (e.g. speech blocks, changes in pronunciation, repetitions, \ldots).
Annotating audio to identify some types
of problems is not the usual way of working in the field of speech therapy,
which has unfortunately led to a shortage of linguistic resources that enable
training automatic evaluation systems or automatic identification of language
pathology.

In this paper we describe the annotation of a speech corpus called \PRAUTOCAL \citep{escudero2022prautocal}, 
which consists of a large number of utterances from individuals with
Down syndrome (DS).
The corpus was recorded using an educational game for training the speech
of individuals with DS.
People with DS have special characteristics that make the production of
quality speech a challenge for them. When the game offers exercises to
practice pronunciation, we must provide as informed and comprehensive
feedback as possible. In order to train a system to perform this task
automatically, it is necessary to have an annotated corpus.
In this work, we present the annotation rubric and the initial results.

The corpus was annotated by an experienced speech therapist, possessing
several years of expertise. This annotation encompasses details regarding
phonetic pronunciation, fluency, and prosodic quality. Initial experiments
were conducted using this annotated data
to predict phonetic quality and fluency errors.

The structure of the paper is as follows.
First, we present the state of the art regarding other rubrics used in other domains.
Next, we describe the rubric and the process that has led to its development.
Then, we present the initial results of using this rubric for the identification of features that indicate voice pathologies.
We conclude with a discussion, conclusions, and future work.

\section{State of the Art}
\label{sec:state}

We find references of rubrics for evaluating or characterizing speech pronunciation in the domains of second language learning and reading proficiency. 
The rubrics within the domain of pathological speech are also becoming 
common.

When it comes to evaluating reading proficiency, \citet{godde2020review} offer a valuable distinction between unidimensional and multidimensional scales. Unidimensional scales assign a single value to encompass various aspects such as decoding, phrasing intonation, and expressivity, as exemplified in works by \citet{pinnell1995listening} and \citet{zutell1991training}. These scales were primarily developed for assessing students by teachers, with the initial three levels emphasizing grouping skills and expressivity introduced at the fourth level. However, the term ``expressivity'' remains somewhat ambiguous, open to personal interpretation by assessors. To enhance the assessment, \citet{zutell1991training} introduced a multidimensional scale designed for teacher use. Subsequently, \citet{rasinski2005reading} further refined the assessment process by differentiating four essential features: pace, smoothness, phrasing, and expression, each rated on a 4-point scale ranging from poor to correct performance. More recently, \citet{haskins2014toward} have suggested focusing on the categories of phrasing and expression for a more comprehensive evaluation of reading proficiency.  

In the domain of L2 pronunciation training, \citet{isaacs2017pronunciation} show the timeline of how state of the art has been evolving. The definition of a standard has driven this timeline as the criteria for assessing pronunciation quality has been controversial (e.g., native like accuracy vs. intelligible/comprehensible speech). The appearing of automatic systems for the evaluation of pronunciation quality put the focus on the criteria used for labeling the training corpus of the automatic systems.   This kind of systems are normally trained
with recordings of manually assessed audios recorded during real tests \citep{litman2018speech}. The rubrics in these cases are limited to setting a binary level or a grade in a scale like in the IELTS test \citep{suzukida2022second}. In spite there exist references that establish pronunciation requirements depending on the evaluated level (for example, Spanish Cervantes Institute recommends in \citealp{santa2007plan} to
consider intonation, stress, phonemes production, \ldots), in practice, students are graded with single scales.

Concerning pathological speech, \citet{demenko2004suprasegmental} use the GRBAS scale to annotate a speech corpora and correlate the scores with acoustic parameters -- global Grade of dysphonia (G), Roughness (R), Breathiness (B), Asthenicity (A), and Strain (S). \citet{saenz2006automatic} use this scale for training automatic evaluation systems. In \citet{bayerl2022ksof}  a corpus with stuttered speech was annotated with the identification of blocks and repetitions. 
In \citet{werle2022professors} stuttering is cross-analyzed with the perception of good communication skills.
In \citet{escudero2022prautocal} 
the quality of oral productions of individuals with DS was annotated,
and a comprehensive review of the currently available corpora was presented.

When several evaluators assess the same pronunciation profile, the inter-rater agreement is an important concern due to the need to validate the quality of the information provided by the evaluators. In that cases, a minimum level of agreement must be set for the test to be satisfactory (i.e. kappa-index > 0.80). That level is reached after appropriate group training (e.g., \citealp{moser2014reliability,paige2014interpreting,schwanenflugel2016development,godde2017evaluation}). When pathological speech is analyzed, inter-rater agreements tend to be low (i.e. kappa-index = 0.35 for blocks in \citealp{bayerl2022ksof}), due to the heterogeneity of the sample.

\section{Rubric Description}
\label{sec:rubric}

Rubrics serve as valuable tools for assessment when they are thoughtfully designed, incorporating four essential criteria: validity, reliability, fairness, and efficiency. When a rubric is considered valid and reliable, it is capable of grading the work at hand objectively, minimizing the impact of any potential instructor biases. In this way, it ensures that anyone employing the rubric will arrive at the same grade for the assessed work.

So, we followed a three steps designing approach to the rubric. First, a draft was elaborated by the authors and validated by expert linguists and therapists. Then, it was used to conduct a controlled evaluation campaign in which a reduced number of samples and experts participated. Finally, from the conclusions in this step and after analyzing the main sources for inter-rater disagreement, a final version was delivered and used by an evaluator which had not participated in previous steps.

We have developed a website to support the evaluation process using the rubric. Additionally, a support page has been designed to provide a detailed rubric description, completed with illustrative examples, to aid evaluators in comprehending each aspect and section. 

We have formulated an evaluation rubric that encompasses three dimensions: phonetics, fluency, and prosody. Each of them is presented below.

\subsection{Phonetics}
\label{sec:phonetics}

\paragraph{Identification of phonetic pronunciation errors at segmental level in words.} When evaluating each word in the utterance, it is essential to identify whether the sequence of phonemes of the word correspond to what is expected for this particular word. The evaluator should mark 
articulation errors that may occur:
substitution, omission, distortion, and addition (SODA)
\citep{Sreedevi2022,gallardo1995alteraciones,susanibar2016principios}.

\begin{itemize}
\item
Substitution: one phoneme is substituted for another. For example, 
\textipa{/pwe\textfishhookr te/} instead of \textipa{/pwe\textfishhookr ta/}
(door), where the phoneme \textipa{/a/} is substituted for \textipa{/e/}. 

\item
Omission: a phoneme is deleted. For example,  
\textipa{/pje\dh a/} instead of \textipa{/pje\dh\textfishhookr a/} 
(stone), where \textipa{/\textfishhookr/} is omitted.

\item
Distortion: a phoneme is not replaced by another, but is not articulated according to what is expected. 

\item
Addition: a phoneme is added to the word that does not belong to it, e.g., 
\textipa{/ba\textbeta je\textfishhookr ta/} instead of 
\textipa{/a\textbeta je\textfishhookr ta/} 
(open), where \textipa{/b/} is added in initial position.
\end{itemize}

\paragraph{Qualitative assessment.} The evaluator must assess the overall phonetics quality of the sentence using the following three levels (where ``sporadic'' means that errors occur in 25\% of the words or less within the utterance; and ``frequent'' if they are present in more than 25\% of the words in the utterance):
\begin{enumerate}
\item
Speech with frequent errors or distortions of sounds. 
\item
Errors are sporadic, appearing in some situations but not in others. 
\item
Pronunciation is correct without obvious pronunciation errors. 
\end{enumerate}

\paragraph{Observations.} Optionally, any additional information considered relevant can be entered in a free-text field. For example, if the word is unintelligible or almost unintelligible, or if it is only understood with the aid of the context. 

\subsection{Fluency}

\paragraph{Identification of fluency errors.}
For each category of fluency errors, the evaluator must indicate whether each deviation occurs once, more than once, or not at all. 

\begin{itemize}
\item
Block: An involuntary pause before a word or within a word, due to physiological reasons. Sometimes, a breath can be perceived, but in most cases, there is no breath or it is inaudible. These pauses are not linguistically motivated. 
For example: \textit{``¿Puedo comprar ... una linterna?''} (Can I buy ... a flashlight?).

\item
Prolongation: An involuntary lengthening or prolongation of a syllable or sound. For example: \textit{``Una escalera de cue[eee]rda''} (A str[iiii]ng ladder).

\item
Repetition of sounds/syllables. 
For example: \textit{``[que-que-]quería comprar una''} (I [wan-wan-]wanted to buy one).

\item
Repetition of words/phrases. 
For example: \textit{``¿Sabes dónde vive la [la] [la] señora Luna?''} (Do you know where Mrs. [Mrs.] [Mrs.] Luna lives?). 

\item
Interjection: Filler words, which are used to buy time to find the right words to continue speaking. They are usually employed when the speaker has difficulties to pronounce a specific word, and fillers provide time to think of an alternative word that is easier to pronounce. 
For example: \textit{``¿Estás [eh] [um] enfermo?''} (Are you [ehh] [uhmm] ill?).
\end{itemize}

\paragraph{Qualitative assessment.} The evaluator should determine the overall quality of the fluency of the sentence using the following three levels (where ``sporadic'' means in 25\% of the words of the utterance or less, and ``frequent'' means more than 25\% of the words of the utterance):
 
\begin{enumerate}
\item
Speech with frequent fluency errors. 
\item
Speech with sporadic fluency errors. 
\item
Speech without fluency errors.
\end{enumerate}

\subsection{Prosody}

\paragraph{Identification of atypical prosody.} For each type of deviation in prosody, the evaluator must indicate whether it occurs once, more than once or not at all. 

\begin{itemize}
\item
Errors in lexical accent. It is necessary to observe both the position of the accent in the framework of the word and the phonetic realization of the stressed-unstressed syllable contrast. What is expected is that the speaker appropriately places the lexical accents of the phrase (that is, differentiates between oxytone, paroxytone, and proparoxytone words) and performs them phonetically according to the prototypical acoustic indices in Spanish (stressed syllables are longer and more intense than the unstressed ones). 

Each time that there are changes in the position of the stressed syllable or in its acoustic manifestation, an error is counted.  

\item
Errors in prosodic grouping. It is necessary to observe the maintenance of the form of the words and their grouping.
What is expected is that the speaker appropriately groups the speech continuum and does not pronounce it as if the words appeared in isolation, in a list format. 
That is what we considered deviations on syllabification (dictation intonation):
each syllable receives a lexical accent, and there is no prosodic modeling of the word, making it sound as if the syllables are separated. For example, \textit{``al-cal-de''} (mayor).

\item
Errors in intonational modality. What is expected is that the speaker clearly produces the melodic variations that correspond to each of the sentence meanings (modalities): declarative, interrogative, exclamatory phrases. 

Errors are considered when there is a lack of correspondence of the melodic curve with the expected modality, in the sentence as a whole, or in some parts of it, so that the intonational meaning cannot be adequately interpreted.  For example, an absolute interrogative that is not produced with a rising ending cannot be interpreted as a question. 
\end{itemize}

\paragraph{Qualitative assessment.}
The evaluator has to determine the overall quality of the prosody of the sentence using the following three levels: 

\begin{enumerate}
\item
Flat speech, without variations in tone, intensity or duration throughout the utterance. 
\item
The prosodic contours of the sentences do not follow the expected pattern, but in general, it does not affect comprehension. 
\item
The prosodic contours are varied and help understand the meaning. 
\end{enumerate}

\subsection{Global Observations}

The evaluator can indicate whether any of the following statements are true or not.

\begin{itemize}
\item
Exaggeratedly deep (hoarse) or high-pitched (shrill) tone of voice. The tone is not in the expected range for the speaker's voice type. There is excessive hoarseness or a tendency towards high-pitched tones. 
\item
Decreased voice volume or with exaggerated variations. Overall, in the statement, the voice volume is too high or insufficient, which hinders understanding of the content. 
\end{itemize}

\paragraph{Additional considerations.}
Optionally, any additional global information considered relevant can be entered in a free-text field.

\section{Annotation of the \PRAUTOCAL Corpus}

In this section, we describe how the rubric presented in the previous section was used to label a subset of \PRAUTOCAL corpus.
We begin with a summary description of this corpus, to ease understanding of the rest of the section.

\subsection{Description of the \PRAUTOCAL Corpus}

The \PRAUTOCAL corpus is a corpus of Spanish speakers with Down syndrome from
the northern/central Iberian Peninsula, which allows the analysis of
specific aspects of the speech of individuals with Down syndrome. It also
includes comparable recordings of typically developing (TD) 
users for reference.
So far, the corpus has been used for prosodic studies. 
In the work described in this paper 
we aim to also use the corpus to evaluate
phonetic pronunciation and fluency.
The corpus was
collected in six recording campaigns and contains 90 speakers, with 4,175
audio files and a total audio duration of 13,633.3 seconds. The corpus is
balanced in terms of gender (49 men and 41 women) and speaker type (50
individuals with intellectual disabilities and 40 typically developing
speakers). The age range for both speaker types is also similar, between 13
and 42 years for speakers with DS and between 6 and 68 years for
TD speakers. 
In this paper, we describe the annotation of the corpus 
in three dimensions:
phonetics, fluency and prosodic quality.
The corpus also includes the
transcription of the utterances, which has been used for
automatic phonetic segmentation of the recordings. A detailed description of the
\PRAUTOCAL corpus can be found in 
\citep{escudero2022prautocal}.

\subsection{Annotation Statistics}

\begin{figure*}[t]
\begin{center}
\includegraphics[width=10cm]{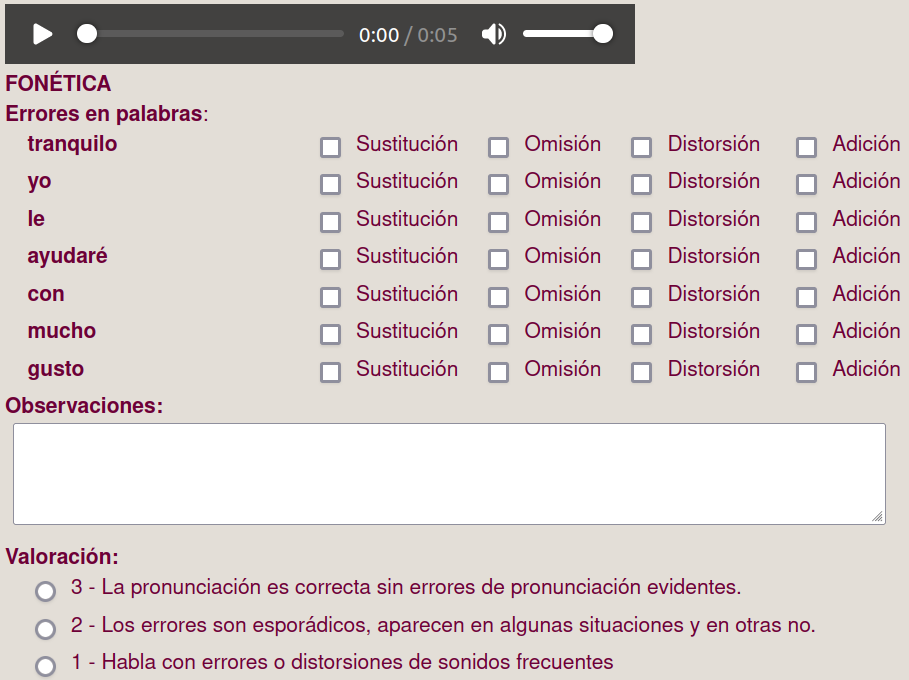} 
\caption{Screenshot of the
phonetic section of the evaluation website. 
The sentence \textit{``Tranquilo. Yo le ayudaré con mucho gusto.''}
(Don't worry. I will be happy to help you.)
has to be evaluated.}
\label{fig:web}
\end{center}
\end{figure*}

A speech therapist with several years of experience
has annotated part of the corpus using the rubric described
in section \ref{sec:rubric}. A total of
1,160 utterances (6,845 words in the phonetic part)
have been annotated. 
The annotation was made using a web page in which the speech therapist could listen to the phrase to be tagged as many times as she wanted and in which 
the errors and the assessment had to be indicated.
Figure \ref{fig:web} shows a screenshot of the phonetic section of the evaluation website.

\begin{figure}[t]
\begin{center}
\includegraphics[width=\linewidth]{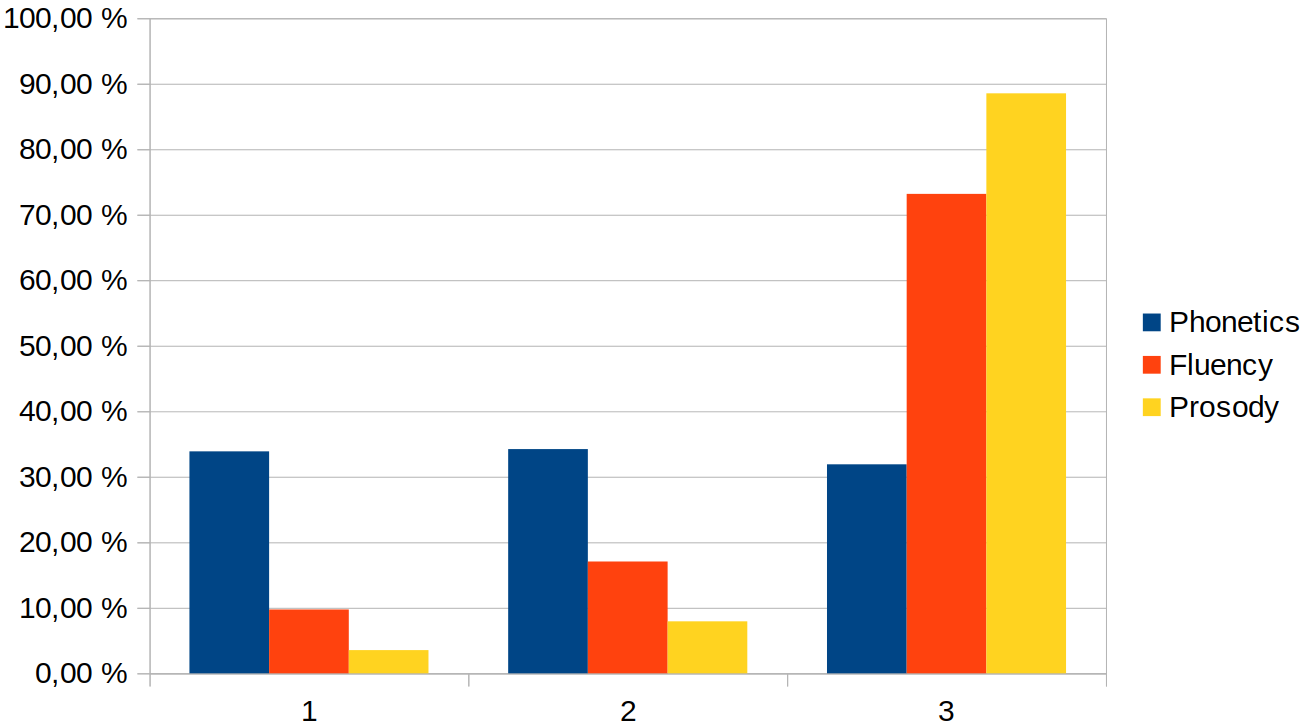} 
\caption{Score distribution for phonetics, 
fluency and prosody.}
\label{fig:dist}
\end{center}
\end{figure}

Figure \ref{fig:dist} shows the distribution of the 
assessment for phonetics, fluency and prosody. 
The distribution of scores for phonetics is quite balanced. However, the distribution of scores 
for fluency and especially for prosody is clearly unbalanced.
The errors
are shown in tables 
\ref{tab:erroresPal},
\ref{tab:erroresFluency} and
\ref{tab:erroresProsody}.

\begin{table}[t]
\centering
\begin{tabular}{lr}
\hline
Substitution & 311 (4.5\%) \\
Omission & 624 (9.1\%) \\
Distortion & 1,799 (26.3\%) \\
Addition & 158 (2.3\%) \\
\hline
\end{tabular}
\caption{\label{tab:erroresPal}Errors in the phonetic
part. The errors are marked at word level.}
\end{table}

\begin{table}[t]
\centering
\begin{tabular}{lrrr}
& \textbf{None} & \textbf{One} & \textbf{Multiple}\\
\hline
Blocks        &  876  &  197  & 87 \\
Prolongations  &  1,029 &  81   & 50 \\
Sound repetitions     &  986  &  127  & 47 \\
Word repetitions    &  1,022 &  89  & 49\\
Interjections  &  1,112 &  40   &  8 \\
\hline
\end{tabular}
\caption{\label{tab:erroresFluency}Errors in the fluency
part.}
\end{table}

\begin{table}[t]
\centering
\begin{tabular}{lrrr}
& \textbf{None} & \textbf{One} & \textbf{Multiple}\\
\hline
Accent        &  1,142  &  16  &  2 \\
Grouping    &  1,069  &  33  &  58 \\
Modality     &  1,094  &  38  &  28 \\
\hline
\end{tabular}
\caption{\label{tab:erroresProsody}Errors in the 
prosodic part.}
\end{table}

\section{Experiments}

In this section we describe the experiments used to compute baseline classification results.
These preliminary experiments 
use common techniques for phonetic assessment and fluency
evaluation.

\subsection{Phonetic Experiments}

In this experiment we have used
Goodness of Pronunciation (GoP) \citep{witt2000phone} for
automatic assessment of phonetic quality.
GoP is a measure of the degree of similarity between produced 
and canonical pronunciation of phonemes.
Although the GoP method is commonly utilized to assess the pronunciation of
non-native (L2) speech, certain studies have also verified its applicability
in evaluating speech disorders
\citep{pellegrini14_interspeech,yeo23_interspeech}.

There are several GoP methods described in the bibliography. 
The first method described was GMM-GoP \citep{witt2000phone}, which,
for a phone $p$ is 
defined as an averaged log probability across the phone duration:

\begin{equation}
\textrm{GMM-GoP} (p) = \frac{1}{|F|} \sum_{f \in F} log \frac
{e^{L^f(p|f)}}
{\sum_{q \in Q} e^{L^f(q|f)}}
\end{equation}

\noindent
where the duration of phone $p$ in frames is $F$;
$P^f(p|f)$ is the frame-wise phone probability and
$L^f(p|f)$ its logits; $Q$ is the total phone set.

With the development of neural networks (NNs), variants of
GoP which employ probabilities from the state-of-the-art neural
networks have been suggested \citep{hu2015improved}:

\begin{equation}
\textrm{NN-GoP} (p) = 
\log
\bar{P}(p|F) 
- \max_{q \in Q}
\log
\bar{P}(q|F) 
\end{equation}

\begin{equation}
\bar{P}(p|F) 
= \frac{1}{|F|} \sum_{f \in F} P^f(p|f)
\end{equation}

Finally, DNN-GoP \citep{hu2015improved} normalizes the phone probability
with the phone prior:

\begin{equation}
\textrm{DNN-GoP} (p) = 
\frac{ \bar{P}(p|F) }{ P(p) }
\end{equation}

In order to calculate the GoP measures, 
we employ a self-supervised learning approach to extract posterior
probabilities from the widely adopted cross-lingual wav2vec 2.0 XLS-R model
\citep{babu22_interspeech}, 
in line with recent research \citep{xu2021explore,yeo23_interspeech}.
The fine-tuning process
uses the 
Common Phone dataset 
\citeplanguageresource{klumpp_2022_5846137}
to train a linear phone prediction head 
on top of the wav2vec
model. Notably, this linear phone prediction head is incorporated above the
convolutional layer rather than the transformer layer. This design choice
serves to 
reduce the computational complexity of the model
while preserving
important phonetic characteristics in the convolutional features. The
optimization is performed using the AdamW optimizer
\citep{Loshchilov2019DecoupledWD}, employing a default
learning rate of 0.001, and this process is repeated for four epochs.

The acoustic model has been trained on a collection of speech samples of typical healthy speakers
drawn from the Common Phone dataset.
This dataset was specifically chosen for
its comprehensive coverage of phonemes and detailed phonetic annotations,
making it a prime choice for encompassing a broad spectrum of Spanish
phonemes. The Common Phone dataset is noteworthy for its gender balance and
multilingual content, spanning six different languages. With over 11,000
speakers contributing to it, the dataset boasts approximately 116 hours of
recorded speech.

We evaluated the efficacy of the model using the \PRAUTOCAL corpus.
First,
in order to obtain the segmentation, Montreal Forced Aligner (MFA) 
\citep{mcauliffe17_interspeech} 
is employed to extract phoneme-level alignments
in the \PRAUTOCAL corpus.
Then, 
we calculated the average GoP score for each utterance and determined its
correlation with the intelligibility score.
We used the Kendall Rank Coefficient $\tau$
to measure the correlation between 
the average GoP scores and the phonetic quality assessments.
Two different configurations were used:

\begin{itemize}
\item 3-level. We used 1,160 utterances from people with DS, evaluated
with the 3-level assessment (1, 2 or 3) as described in section
\ref{sec:phonetics}.
\item 4-level. We added 380 
randomly chosen utterances of 
typically developing (TD) speakers from the \PRAUTOCAL corpus.
As they are supposed to be correctly pronounced utterances,
the assessment was set to 4.
\end{itemize}

\begin{figure}[t]
\begin{center}
\includegraphics[width=\linewidth]{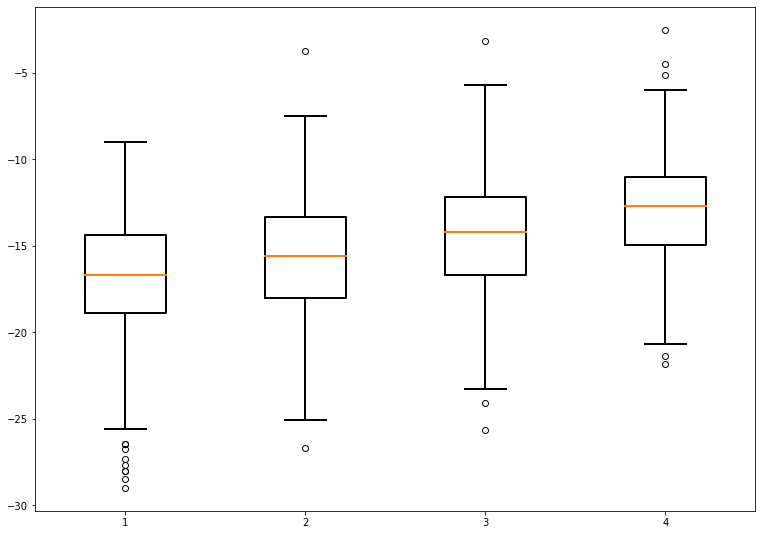} 
\caption{Distribution of the GMM-GoP for the four
different score levels.}
\label{fig:gmm}
\end{center}
\end{figure}

The achieved results are shown in table \ref{gop-results},
highlighting the performance of different configurations.
In the 3-level setup, the best outcome is achieved by 
GMM-GoP, with a score of 0.2089. 
However, when employing the 4-level configuration, 
the best result is obtained by DNN-GoP, attaining a score of 0.3544.
Notably, the 4-level configuration consistently outperforms the 3-level counterpart in all scenarios. 
Figure \ref{fig:gmm} 
offers an insight into the distribution of GMM-GoP measurements
for scores 4, 3, 2 and 1 
(similar graphics are observed for NN-GoP and for DNN-GoP).
The highest GoP values are obtained for score 4 (TD speakers),
as their speech is expected to be very similar to the
canonical pronunciation of phonemes,
while the lowest values are associated with score 1. 
Scores 2 and 3 exhibit intermediate values between score 1 and score 4. 
In summary, the results align with anticipated trends, 
despite the somewhat modest correlation values.

\begin{table}[t]
\centering
\begin{tabular}{lll}
& \textbf{3-level} & \textbf{4-level}\\
\hline
GMM-GoP & 0.2089 & 0.3109 \\
NN-GoP  & 0.1852 & 0.3056 \\
DNN-GoP & 0.1344 & 0.3544 \\
\hline
\end{tabular}
\caption{\label{gop-results} Kendall's rank coefficient 
between GoP measures and phonetic evaluation scores of the rubric.}
\end{table}

\subsection{Stuttering Experiments}

To evaluate the annotations for stuttering made on the sentences of the corpus, we used a similar approach to \citet{bayerl2022detecting}. As a reference corpus, we utilized the KSoF corpus \citeplanguageresource{ksof22}, which is a corpus containing 5,597 3-second recordings obtained from stuttering therapy sessions. These recordings were labeled by three annotators for five types of disfluencies: blocks, prolongations, sound repetitions, word repetitions, and interjections, in addition to other additional labels that we have not considered in this study. The distribution of labels in this corpus is highly unbalanced.

As feature extractor, we used the base model of wav2vec \citep{baevski2020wav2vec}, which is a model trained in an unsupervised manner with 960 hours of unlabeled speech data from the LibriSpeech corpus 
\citeplanguageresource{librispeech2015}.
This model was adapted for automatic speech recognition using transcriptions of the same audio. For each labeled audio, we extracted 768-dimensional speech representations every 20ms, and the mean was calculated to obtain representations for the entire audio. 
We used SVMs as classifiers because they have demonstrated effectiveness in delivering robust results even when working with a limited set of samples.
The classification is a binary tasks of one specific disfluency against all other samples.

To obtain the optimal hyper-parameters of the classifier for each type of disfluency, we conducted classification using the embedding extracted from each of the layers provided by the model, employing principal component analysis (PCA) to obtain features that explain a minimum of 0.9 variance. The kernel parameters were selected from \{0.1, 0.01, 0.001, 0.0001, 0.00001\}, and the penalty parameter of error C was chosen from \{1, 10, 100, 1000\}. From the results obtained, we selected the parameters and layer that achieved the best results (F1 score) for each type of disfluency. Finally, an SVM classifier was trained using all the samples with the parameters obtained in the previous phase.

The classifier trained on the KSoF corpus was used to classify samples from our corpus and the results are shown in Table \ref{tab:stuttering}. As can be seen in Table \ref{tab:erroresFluency}, the labels used for the fluency evaluation are similar to those used in the KSoF corpus for the purpose of comparison. However, it is important to note that both the evaluators and their interpretation of the evaluation criteria may differ. As a result, we cannot establish an exact correspondence between the labels in both corpora, even if the disfluency to be detected is the same. This variation can impact on the classification results. Additionally, just like in the KSoF corpus, the labels for different disfluencies are highly unbalanced. In the \PRAUTOCAL corpus, the identification of disfluencies is not binary, as the evaluator has three levels of assessment for each type of disfluency: no disfluency, one disfluency, or two or more disfluencies. To convert this assessment into a binary format, it is considered that there is disfluency if the label indicates one or more disfluencies.

For each of the labels, the F1 score for disfluency identification/non-identification has been calculated. For blocks, an F1 score of 0.31/0.67 is obtained; for prolongations, an F1 score of 0.22/0.37; for sound repetitions, an F1 score of 0.19/0.88; for word repetitions, an F1 score of 0/0.93; and for interjections, an F1 score of 0.08/0.68 is obtained.

\begin{table}[t]
\centering
\begin{tabular}{lrr}
& \textbf{KSoF} & \textbf{\PRAUTOCAL(id/nid)} \\
\hline
Blocks  & 0.56 & 0.31 / 0.67    \\
Prolongations & 0.58 & 0.22 / 0.37   \\
Sound repetitions & 0.36 & 0.19 / 0.88    \\
Word repetitions  & 0.44 & 0 / 0.93    \\
Interjections & 0.16 & 0.08 / 0.68    \\
\hline
\end{tabular}
\caption{F1 score of each disfluency label. KSoF column shows the best result in the optimization process (disfluency detection). \PRAUTOCAL column shows the results obtained with the classifier trained with the KSoF samples and tested with the \PRAUTOCAL corpus. Id means disfluency detection and nId means no disfluency detection}
\label{tab:stuttering}
\end{table}

\section{Discussion}

To demonstrate the efficacy of our rubric, a series of experiments were conducted. Within the realm of phonetic experimentation, we presented baseline machine learning trials, yielding results indicating a moderate correlation between the GoP values and assessment scores. This outcome was achieved through a straightforward approach of averaging GoP across all phonemes within an utterance. While this serves as an initial step for future research, there is potential to enhance results through the implementation of novel classification techniques. Moreover, it is imperative to conduct an in-depth examination of GoP values for each individual phoneme. Individuals with Down syndrome often face greater challenges in articulating specific phonemes while displaying relative proficiency in others. For instance, in the context of the Spanish language, individuals with DS encounter heightened difficulty in acquiring the /rr/, /r/, and /z/ phonemes \citep{marin2014comparacion}.

The results in Table \ref{tab:stuttering} show the difficulty of evaluating different disfluencies using automatic classifiers. The best results are obtained for blocks and prolongations, although they are still far from the results achieved in the KSoF corpus. Nonetheless, F1 values for the absence of disfluency are higher for all types of disfluencies. This is particularly important during therapy sessions, as an inaccurate classification of disfluency can lead to patient frustration. Labeling the corpus, even when considering the same disfluencies, is a complex process because the interpretation of different disfluencies can vary among evaluators. In the KSoF corpus used to train the classifier, there is not very high agreement among evaluators \citep{bayerl2022ksof}, indicating the difficulty of applying the same criteria when evaluating speech disfluencies.

\section{Conclusions}

This paper introduces a rubric for identifying errors in 
phonetics, fluency and prosody
in the speech of individuals with Down syndrome. Additionally, it presents two preliminary experiments aimed at assessing the quality of these annotations in automatic classification systems. While the initial results are subject to improvement, having a corpus annotated for individuals with Down syndrome across these three aspects of speech opens up the potential to enhance these outcomes, ultimately paving the way for the development of automatic systems that can be beneficial in speech therapy for individuals with Down syndrome.

Other aspects related to the collection of the corpus, such as the recording process and the phrases to be evaluated, can also influence labeling and the subsequent automatic evaluation of these disfluencies. As future work, it is important to consider the inclusion of new evaluators to improve the labeling. In addition, fine-tuning stuttering models can be applied, along with the implementation of other deep learning techniques that have the potential to improve the results.

\section{Acknowledgements}

This work was carried out in the Project \ProjectCode that was
supported by \FundingAgency.

\section{Bibliographical References}\label{sec:reference}

\bibliographystyle{lrec-coling2024-natbib}
\bibliography{bibliografia}

\section{Language Resource References}
\label{lr:ref}
\bibliographystylelanguageresource{lrec-coling2024-natbib}
\bibliographylanguageresource{languageresource}
\end{document}